\documentclass{article}
\usepackage{spconf,amsmath,epsfig}
\usepackage{multirow}
\usepackage{hhline}
\usepackage{colortbl}
\usepackage{float}
\usepackage{booktabs}
\usepackage{amsmath, amssymb}
\usepackage{bbding}
\usepackage{utfsym}
\let\OLDthebibliography\thebibliography
\renewcommand\thebibliography[1]{
  \OLDthebibliography{#1}
  \setlength{\parskip}{0pt}
  \setlength{\itemsep}{0pt plus 0.3ex}
}
\usepackage{enumitem}
\usepackage{hyperref} 
\hypersetup{
colorlinks=true,
linkcolor=black
}
\begin{document}\sloppy
\def\x{{\mathbf x}}
\def\L{{\cal L}}

\title{HDBN: A Novel Hybrid Dual-branch Network for Robust Skeleton-based Action Recognition}
\name{Jinfu Liu$^*$$^1$, Baiqiao Yin$^*$$^1$, Jiaying Lin$^1$, Jiajun Wen$^1$, Yue Li$^1$, Mengyuan Liu$^\dagger$$^2$}
\address{$^1$School of Intelligent Systems Engineering, Sun Yat-sen University \\
$^2$National Key Laboratory of General Artificial Intelligence, Peking University, Shenzhen Graduate School\\
$^*$ means co-first authors with equal contributions. \\
Corresponding author is Mengyuan Liu (e-mail: liumengyuan@pku.edu.cn). }

\maketitle
\begin{abstract}
Skeleton-based action recognition has gained considerable traction thanks to its utilization of succinct and robust skeletal representations. Nonetheless, current methodologies often lean towards utilizing a solitary backbone to model skeleton modality, which can be limited by inherent flaws in the network backbone. To address this and fully leverage the complementary characteristics of various network architectures, we propose a novel Hybrid Dual-Branch Network (HDBN) for robust skeleton-based action recognition, which benefits from the graph convolutional network's proficiency in handling graph-structured data and the powerful modeling capabilities of Transformers for global information. In detail, our proposed HDBN is divided into two trunk branches: MixGCN and MixFormer. The two branches utilize GCNs and Transformers to model both 2D and 3D skeletal modalities respectively. Our proposed HDBN emerged as one of the top solutions in the Multi-Modal Video Reasoning and Analyzing Competition (MMVRAC) of 2024 ICME Grand Challenge, achieving accuracies of 47.95$\%$ and 75.36$\%$ on two benchmarks of the UAV-Human dataset by outperforming most existing methods. Our code will be publicly available at: \href{https://github.com/liujf69/ICMEW2024-Track10}{\textcolor{black}{https://github.com/liujf69/ICMEW2024-Track10}}.
\end{abstract}

\begin{figure*}[t] 
  \centering
  \includegraphics[width=0.9\linewidth]{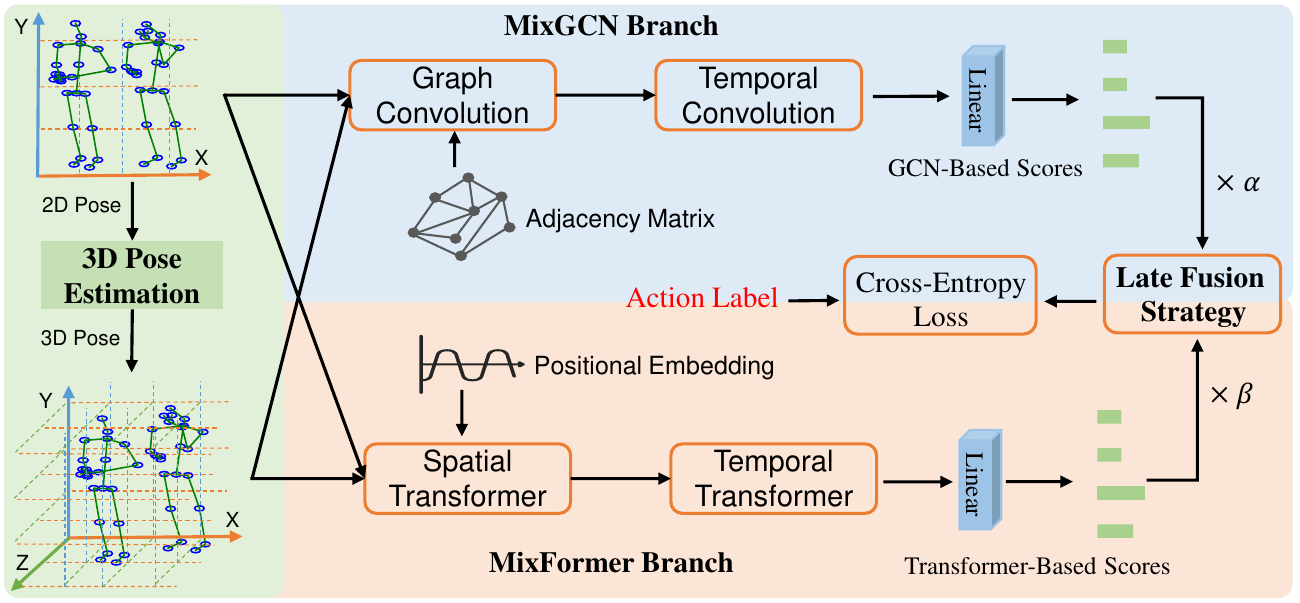}
  \caption{Framework of our proposed Hybrid Dual-branch Network.}
  \label{framework}
\end{figure*}

\section{Introduction}
\label{sec:intro}
Human action recognition is an important task in video understanding, which also has applications in research fields including human-robot interaction and security surveillance. Due to the robustness against environmental interference and the effective representation of body movements, the skeleton has become the mainstream modality used in human action recognition. In previous studies \cite{LIU2017346, 8306456, yan2018spatial, shi2019two, shiftgcn2020, CTR-GCN2021, TDGCN}, skeletons were commonly represented in the form of pseudo-images and topological graphs, while being modeled using classic neural networks simultaneously.

Initially, skeletons were transformed into 2D pseudo-images \cite{LIU2017346, 8306456} and feature extraction was performed using convolutional neural networks (CNNs), which effectively utilized the modeling capability of convolutional layers for Euclidean data. However, these methods tended to overlook the connections between skeleton joints. Indeed, the skeleton modality can be perceived as an inherent topological graph, with the graph vertices and edges symbolizing the joints and bones of the human body respectively. Therefore, graph convolutional networks (GCNs) are used to model skeletons in many studies \cite{yan2018spatial, shi2019two, shiftgcn2020, CTR-GCN2021, TDGCN, Chi_2022_CVPR, bai2022hierarchical, liu2023novel}. However, GCNs are often limited by the design of the adjacency matrix, making it challenging to migrate to skeleton inputs with different numbers of joints. Currently, due to the powerful abstraction and modeling capabilities of Transformers for global information, some studies have applied Transformers \cite{wen2023interactive, xin2023skeleton} in action recognition. Similarly, Transformers face the issue of overlooking the design of adjacency matrices for skeletal joints, resulting in recognition accuracy inferior to state-of-the-art GCNs.

In reality, CNNs, GCNs and Transformers each have their own unique advantages and disadvantages when processing skeleton modality, leading to poor performance in fine-grained recognition of certain human actions. A good approach to address the above issues is to simultaneously use multiple network structures, achieving robust skeleton-based action recognition through the complementarity between networks. Although there have been existing solutions \cite{das2020vpn, ding2023integrating, EPPNet, MMNet} that simultaneously use CNN and GCN networks, they often apply to scenarios where both RGB image modality and skeleton modality are used as inputs.

Therefore, in order to better handle scenarios with only skeleton inputs and leverage the complementarity between different backbone networks, we propose a novel Hybrid Dual-Branch Network (HDBN) for robust skeleton-based action recognition. Specifically, our HDBN consists of two trunk branches called MixGCN branch and MixFormer branch, which is shown in Fig.\ref{framework}. In the MixGCN branch, we utilize GCNs to separately process 2D and 3D skeleton inputs, and employ a late-fusion strategy to aggregate the classification results from different skeleton input. Similarly, in the Mixformer branch, we employ Transformer to model the skeleton inputs, fully harnessing Transformer's abstraction capability for global information. By leveraging the proposed HDBN, we effectively integrate GCNs and TransFormers to achieve better human action recognition.

Our contributions are summarized as follows:
\setlist{nolistsep}
\begin{itemize}[noitemsep, leftmargin=*]
    \item We advocate for utilizing different network structures to achieve robust skeleton-based action recognition, fully leveraging the structural complementarity between different backbones.
    \item We propose a new dual-branch framework called Hybrid Dual-Branch Network (HDBN), which effectively combines GCNs and Transformers. In detail, the skeleton data are inputted to GCN and Transformer backbones for modeling high-level features, which will be effectively combined through a late fusion strategy for more robust skeleton-based action recognition.
    \item Extensive experiments on benchmark UAV-Human dataset verify the effectiveness of our HDBN. In this large-scale action recognitio datasets, our model outperforms most existing action recognition methods.
\end{itemize}

\section{Related Work}
\subsection{CNN-Based and RNN-Based Methods}
In the realm of skeleton-based action recognition, deep neural network (DNN) approaches predominantly leverage Recurrent Neural Networks (RNNs) and Convolutional Neural Networks (CNNs). RNNs typically incorporate outputs from previous time steps as inputs for the current time step, forming a recursive connection structure that excels in handling sequential data. On the other hand, CNNs harness the unique advantages inherent in processing Euclidean data by often transforming skeleton sequences into 2D pseudo-images. Several prior studies have harnessed the power of RNNs \cite{8226767, 8322061, liu2016spatio} and CNNs \cite{LIU2017346, 8306456} to attain impactful outcomes in action recognition based on skeletal data. However, neither RNN-based nor CNN-based methods have been able to fully capture the intricate dependencies among human joints, thereby limiting advancements in recognition accuracy.

\subsection{GCN-Based Methods}
GCNs represent a class of networks adept at efficiently handling structured data, particularly graph data. In the domain of skeleton-based action recognition, researchers \cite{yan2018spatial, shi2019two, shiftgcn2020, CTR-GCN2021, TDGCN} have harnessed the power of GCNs with promising results. For example, Yan et al. \cite{yan2018spatial} introduced the Spatial-Temporal Graph Convolutional Network (ST-GCN), conceptualizing joints as vertices and predefining edges to capture dependencies informed by prior knowledge. Liu et al. \cite{TDGCN} introduced the Temporal Decoupling Graph Convolutional Network (TD-GCN), innovatively leveraging a temporal-dependent adjacency matrix to extract high-level features from skeleton data with remarkable efficacy. Nevertheless, the efficacy of GCNs is frequently constrained by the fixed design of the adjacency matrix, posing challenges when adapting to skeleton inputs with varying numbers of joints.

\subsection{Transformer-Based Methods}
Transformers inherently excel at capturing and processing global information, which is helpful for improving the classification capabilities in skeleton-based action recognition. Numerous research endeavors have been dedicated to optimizing transformers specifically for action recognition tasks. For instance, to address problems that previous methods are hard to capture interactive relations, Wen et al. \cite{wen2023interactive} proposed an Interactive Spatiotemporal Token Attention Network (ISTANet) based on the Transformer, which simultaneously model spatial, temporal, and interactive relations. To address problems that current skeleton transformer depends on the self-attention mechanism of the complete channel of the global joint, ignoring the highly discriminative differential correlation within the channel, Xin et al. \cite{xin2023skeleton} proposed the Skeleton MixFormer, which effectively represent the physical correlations and temporal interactivity of the compact skeleton data. Nevertheless, Transformer-based approaches often encounter challenges in adequately designing adjacency matrices for skeletal joints, leading to recognition accuracy that falls short of cutting-edge Graph Convolutional Networks.

\subsection{Hybrid Dual-Branch Methods}
Due to the inherent limitations of a single backbone network, it struggles to fully extract and model information from the skeleton modality. Consequently, some research efforts \cite{das2020vpn, ding2023integrating, EPPNet, MMNet} are dedicated to employing multiple backbone networks (such as the combination of CNNs and GCNs) to achieve more robust action recognition. For instance, Liu et al. \cite{EPPNet} utilize both CNN and GCN backbones to process human parsing and human pose modalities separately, which implement a late fusion strategy to combine features from both modalities. Das et al. \cite{das2020vpn} propose the Video-Pose Network (VPN), which employs both CNN and GCN to model RGB and skeletal modalities, enabling the learning of enhanced spatiotemporal features. However, the aforementioned hybrid dual-branch methods tend to rely on both CNNs and GCNs and require input from two different modalities, which is not conducive to modeling situations where only skeleton is available. Distinct from all the methods mentioned above, our proposed HDBN is a novel dual-branch scheme that combines the state-of-the-art GCN and Transformer backbones for robust skeleton-based action recognition.
\begin{center}
\begin{table*}[t]
\footnotesize
\caption{Accuracy comparison with state-of-the-art methods on UAV-Human dataset. Symbol $\star$ represents the result reproduced in our work.}
\begin{tabular*}{\textwidth}{@{\extracolsep\fill}llll@{}}
\toprule
\textbf{Method} & \textbf{Source} & \textbf{CSv1 ($\%$)} & \textbf{CSv2 ($\%$)} \\
\midrule
DGNN \cite{shi2019skeleton} & CVPR'19 & 29.90 & - \\
ST-GCN \cite{yan2018spatial} & AAAI'18 & 30.25 & 56.14 \\
2s-AGCN \cite{shi2019two} & CVPR'19 & 34.84 & 66.68 \\
HARD-Net \cite{li2020hard} & ECCV'20 & 36.97 & - \\
Shift-GCN \cite{shiftgcn2020} & CVPR'20 & 37.98 & 67.04 \\
FR-AGCN \cite{hu2022forward} & Neuroc'22 & 43.98 & 69.50 \\
CTR-GCN$^\star$ \cite{CTR-GCN2021} & ICCV'21 & 45.60 & 72.50 \\
ACFL-CTR \cite{wang2022skeleton} & ACMMM'22 & 45.30 & - \\
TD-GCN$^\star$ \cite{TDGCN} & TMM'24 & 45.43 & 72.86 \\
Skeleton MixFormer$^\star$ \cite{xin2023skeleton} & ACMMM'23 & 47.23 & 73.47 \\
\midrule
\textbf{MixGCN} (Ours) & - & \textbf{46.73} & \textbf{74.06} \\
\textbf{MixFormer} (Ours) & - & \textbf{47.23} & \textbf{73.47} \\
\textbf{HDBN} (Ours) & - & \textbf{47.95} & \textbf{75.36} \\
\bottomrule
\label{tab:sota}
\end{tabular*}
\end{table*}
\end{center}
\begin{table}[t]
\footnotesize
\caption{Accuracy of different 2D skeleton modalities on UAV-Human dataset when using CTR-GCN as the backbone.}
\centering
\begin{tabular}{cccccc}
\hline
\multicolumn{4}{c}{\textbf{Modality}} & \multicolumn{2}{c}{\textbf{UAV-Human} ($\%$)} \\\hline
J & B & JM & BM & \textbf{CSv1} & \textbf{CSv2} \\\hline
\Checkmark & & & & 43.52 & 69.00  \\
& \Checkmark & & & 43.32 & 68.68  \\
& & \Checkmark & & 36.25 & 57.93  \\
& & & \Checkmark & 35.86 & 58.45  \\\hline
\Checkmark & \Checkmark & \Checkmark & \Checkmark & 45.60 & 72.50\\\hline
\end{tabular}
\label{tab:CTR-GCN-2D}
\end{table}
\begin{table}[t]
\footnotesize
\caption{Accuracy of different 3D skeleton modalities on UAV-Human dataset when using CTR-GCN as the backbone.}
\centering
\begin{tabular}{cccccc}
\hline
\multicolumn{4}{c}{\textbf{Modality}} & \multicolumn{2}{c}{\textbf{UAV-Human} ($\%$)} \\\hline
J & B & JM & BM & \textbf{CSv1} & \textbf{CSv2} \\\hline
\Checkmark & & & & 35.14 & 64.60  \\
& \Checkmark & & & 35.66 & 63.25  \\
& & \Checkmark & & 31.08 & 55.80  \\
& & & \Checkmark & 30.89 & 54.67  \\\hline
\Checkmark & \Checkmark & \Checkmark & \Checkmark & 38.85 & 67.87\\\hline
\end{tabular}
\label{tab:CTR-GCN-3D}
\end{table}
\begin{table}[t]
\footnotesize
\caption{Accuracy of different 2D skeleton modalities on UAV-Human dataset when using TD-GCN as the backbone.}
\centering
\begin{tabular}{cccccc}
\hline
\multicolumn{4}{c}{\textbf{Modality}} & \multicolumn{2}{c}{\textbf{UAV-Human} ($\%$)} \\\hline
J & B & JM & BM & \textbf{CSv1} & \textbf{CSv2} \\\hline
\Checkmark & & & & 43.21 & 69.50  \\
& \Checkmark & & & 43.33 & 69.30  \\
& & \Checkmark & & 35.74 & 57.74  \\
& & & \Checkmark & 35.56 & 55.14  \\\hline
\Checkmark & \Checkmark & \Checkmark & \Checkmark & 45.43 & 72.86\\\hline
\end{tabular}
\label{tab:TD-GCN-2D}
\end{table}
\begin{table}[t]
\footnotesize
\caption{Accuracy of different 2D skeleton modalities on UAV-Human dataset when using MST-GCN as the backbone.}
\centering
\begin{tabular}{cccccc}
\hline
\multicolumn{4}{c}{\textbf{Modality}} & \multicolumn{2}{c}{\textbf{UAV-Human} ($\%$)} \\\hline
J & B & JM & BM & \textbf{CSv1} & \textbf{CSv2} \\\hline
\Checkmark & & & & 41.48 & 67.48  \\
& \Checkmark & & & 41.57 & 67.30  \\
& & \Checkmark & & 33.82 & 54.43  \\
& & & \Checkmark & 34.74 & 52.13  \\\hline
\Checkmark & \Checkmark & \Checkmark & \Checkmark & 43.78 & 70.95\\\hline
\end{tabular}
\label{tab:MST-GCN-2D}
\end{table}
\begin{table}[t]
\footnotesize
\caption{Accuracy of different 2D skeleton modalities on UAV-Human dataset when using Skeleton MixFormer as the backbone. The definitions of k2 and k2M can be found in \cite{Chi_2022_CVPR}.}
\centering
\begin{tabular}{cccccccc}
\hline
\multicolumn{6}{c}{\textbf{Modality}} & \multicolumn{2}{c}{\textbf{UAV-Human} ($\%$)} \\\hline
J & B & JM & BM & k2 & k2M & \textbf{CSv1} & \textbf{CSv2} \\\hline
\Checkmark & & & & & & 41.43 & 66.03  \\
& \Checkmark & & & & & 37.40 & 64.89  \\
& & \Checkmark &  & & & 33.41 & 54.58  \\
& & & \Checkmark & & & 30.24 & 52.95  \\
& & & & \Checkmark & & 39.21 & 65.56  \\
& & & & & \Checkmark & 32.60 & 55.01  \\\hline
\Checkmark & \Checkmark & \Checkmark & \Checkmark & \Checkmark & \Checkmark & 47.23 & 73.47 \\\hline
\end{tabular}
\label{tab:Mixformer-2D}
\end{table}
\vspace{-1cm}
\section{Method}
\subsection{Multimodal Data Processing}
In our proposed Hybrid Dual-branch Network (HDBN), we use robust skeleton modality for human action recognition. Conceptually, the sequence of skeletal structure forms a natural topological graph, wherein joints represent graph vertices and bones serve as edges. The graph can be defined as $\textit{\textbf{G}} = \left\{\textit{\textbf{V}}, \textit{\textbf{L}} \right\}$, where $\textit{\textbf{V}} = \left\{\textit{\textbf{v}}_{1}, \textit{\textbf{v}}_{2}, \cdots, \textit{\textbf{v}}_{N} \right\}$ and $\textit{\textbf{L}} = \left\{\textit{\textbf{l}}_{1}, \textit{\textbf{l}}_{2}, \cdots, \textit{\textbf{l}}_{M} \right\}$ represent \textit{N} joints and \textit{M} bones of human body. The joint $\textit{\textbf{v}}_{i}$ is defined as $\left\{\textit{\textbf{x}}_{i}, \textit{\textbf{y}}_{i} \right\}$ in 2D pose data, where $\textit{\textbf{x}}_{i}$ and $\textit{\textbf{y}}_{i}$ locate $\textit{\textbf{v}}_{i}$ in two-dimensional Euclidean space.

Here we define four skeleton modalities, namely joint $(\textit{\textbf{J}})$, bone $(\textit{\textbf{B}})$, joint motion $(\textit{\textbf{JM}})$ and bone motion $(\textit{\textbf{BM}})$. Given two joints data $\textit{\textbf{v}}_{i} = \left\{\textit{\textbf{x}}_{i}, \textit{\textbf{y}}_{i} \right\}$ and $\textit{\textbf{v}}_{j} = \left\{\textit{\textbf{x}}_{j}, \textit{\textbf{y}}_{j} \right\}$, a bone data of the skeleton is defined as $\textit{\textbf{e}}_{\textit{\textbf{v}}_{i}, \textit{\textbf{v}}_{j}} = \left(\textit{\textbf{x}}_{i}-\textit{\textbf{x}}_{j}, \textit{\textbf{y}}_{i}-\textit{\textbf{y}}_{j} \right)$. Given two joints data $\textit{\textbf{v}}_{ti}$, $\textit{\textbf{v}}_{(t+1)i}$ from two consecutive frames, the data of joint motion is defined as $\textit{\textbf{m}}_{ti} = \textit{\textbf{v}}_{(t+1)i} - \textit{\textbf{v}}_{ti}$. Similarly, given two bones data $\textit{\textbf{e}}_{\textit{\textbf{v}}_{(t+1)i},\textit{\textbf{v}}_{(t+1)j}}$, $\textit{\textbf{e}}_{\textit{\textbf{v}}_{ti},\textit{\textbf{v}}_{tj}}$ from two consecutive frames, the data of bone motion is defined as $\textit{\textbf{m}}_{\textit{\textbf{v}}_{ti},\textit{\textbf{v}}_{tj}} = \textit{\textbf{e}}_{\textit{\textbf{v}}_{(t+1)i},\textit{\textbf{v}}_{(t+1)j}} - \textit{\textbf{e}}_{\textit{\textbf{v}}_{ti},\textit{\textbf{v}}_{tj}}$.

Typically, 3D skeleton data provides more information compared to 2D skeleton data, which can compensate for the limitations of 2D data. Therefore, our HDBN will utilize both 2D and 3D skeleton data simultaneously. Specifically, we will employ a 3D pose estimation encoder \cite{Zhu_2023_ICCV} to derive 3D skeleton data from 2D pose data, which can be formulated as:
\begin{equation}
  P_{3D} = Encoder(x_{i}, y_{i})
\label{con:Equation1}
\end{equation}
\subsection{MixGCN Backbone} 
Due to its natural advantages in processing graph-structured data, the MixGCN branch in our proposed HDBN utilizes GCNs to model skeleton data. Actually the Graph Convolutional Networks consist of two key parts: graph convolution module and temporal convolution module. The normal graph convolution can be formulated as:
\begin{equation}
  H^{l+1} = \sigma\left(D^{-\frac{1}{2}}AD^{-\frac{1}{2}}H^{l}W^{l} \right)
  \label{eq:Equation2},
\end{equation}
where $H^{l}$ is the features at layer $\textit{l}$ and $\sigma$ is the activation function. $\textit{D} \in \textit{R}^{N \times N}$ is the degree matrix of $\textit{N}$ joints and $\textit{W}^{\textit{l}}$ is the learnable parameter of the $\textit{l}$-th layer. $\textit{A}$ is the adjacency matrix representing joint connections. Generally, the $\textit{A}$ can be generated by using static and dynamic ways. In the MixGCN branch, we employ the backbone with three types of dynamic adjacency matrices (i.e. TD-GCN \cite{TDGCN}, CTR-GCN \cite{CTR-GCN2021} and MST-GCN \cite{yan2018spatial}) to handle both 2D and 3D skeletal data, fully leveraging the complementarity among different GCNs.

\subsection{MixFormer Backbone}
Since Transformers possess powerful modeling and abstraction capabilities for global information, the MixFormer branch in our HDBN utilizes Transformer as the core backbone to model the skeleton data. Generally, a standard Transformer consists of modules such as positional encoding, self-attention, and feedforward neural networks. We can utilize the Transformer architecture to encode features of input modality and combine specific downstream layers to handle various downstream tasks. A simplified representation of the encoded features is as follows:
\begin{center}
\begin{equation}
\setlength{\abovedisplayskip}{-10pt}
\setlength{\belowdisplayskip}{0pt}
\textit{$\textit{\textbf{F}}$} = \textit{Norm}[\textit{\textbf{X}} + \textit{Attention}(\textit{\textbf{X}}\textit{\textbf{W}}_Q, \textit{\textbf{X}}\textit{\textbf{W}}_K, \textit{\textbf{X}}\textit{\textbf{W}}_V)\textit{\textbf{W}}_O]
\label{con:Equation3}
\end{equation}
\end{center}
Where $\textit{\textbf{W}}_Q$, $\textit{\textbf{W}}_K$, $\textit{\textbf{W}}_V$, and $\textit{\textbf{W}}_O$ represent the query, key, value and output weight matrices respectively.

\begin{figure*}[t] 
  \centering
  \includegraphics[width=0.8\linewidth]{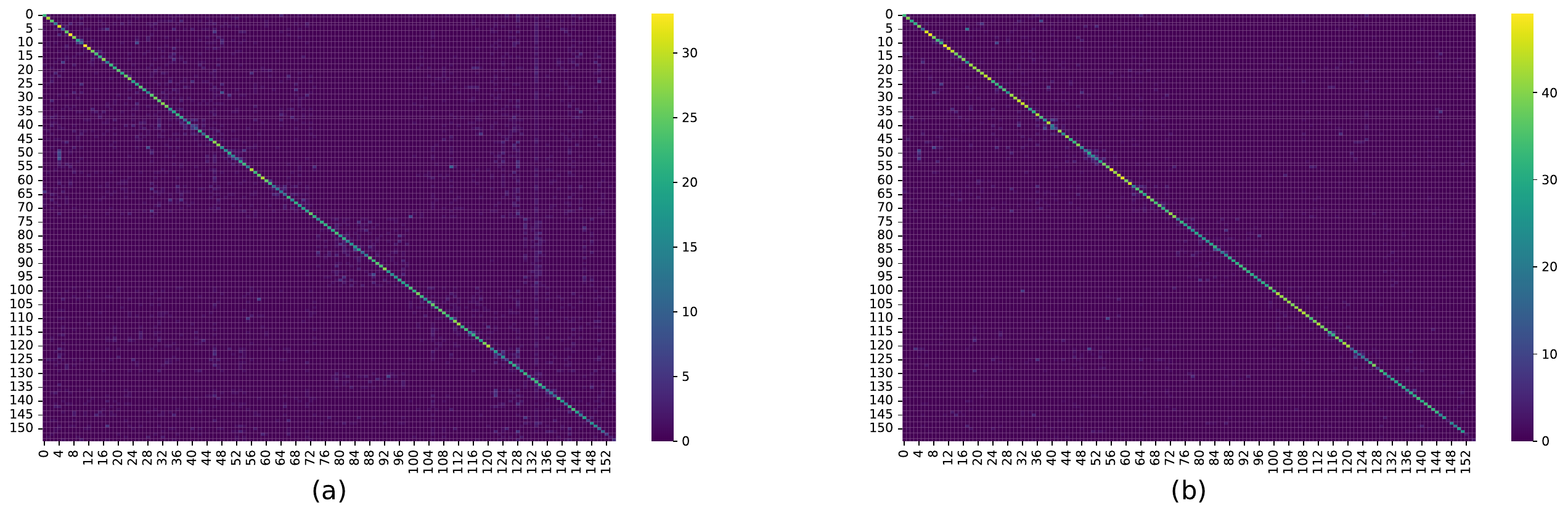}
  \caption{The confusion matrix of UAV-Human. The more yellow the squares on the diagonal, the more accurate the recognition. (a) UAV-Human dataset on the benchmark of CSv1. (b) UAV-Human dataset on the benchmark of CSv2}
  \label{confusion}
\end{figure*}

\subsection{Ensemble}
In our proposed HDBM, we utilize a late-fusion strategy to combine the classification scores from the MixGCN branch and the MixFormer branch. In detail, the late-fusion strategy employs a set of weight parameters to linearly combine the classification scores from the two branches, which is formulated as:
\begin{center}
\begin{equation}
\setlength{\abovedisplayskip}{-10pt}
\setlength{\belowdisplayskip}{0pt}
\textit{$\textit{\textbf{S}}$} = \alpha \cdot \mathcal M_{gcn}(\textit{\textbf{P}}) + \beta \cdot \mathcal M_{former}(\textit{\textbf{P}}),
\label{con:Equation4}
\end{equation}
\end{center}
Where $\mathcal M_{gcn}$ and $\mathcal M_{former}$ represent the MixGCN backbone and MixFormer backbone respectively, while the $\alpha$ and $\beta$ represent the weights for linear fusion. The final result \textit{\textbf{S}} that integrates two branches will be processed by a softmax layer for final robust action recognition.

In our proposed HDBN, we use Cross-Entropy (CE) loss to train the entire network: 
\begin{equation}
  \mathcal L_{cls} = -\sum_{i}^{N}\mathcal Y^{i}log\mathcal S^i,
  \label{eq:Equation5}
\end{equation}
where $\textit N$ is the number of samples in a batch and $\mathcal Y^{i}$ is the one-hot presentation of the true label about action sample $\textit i$. The $\mathcal S^i$ is the true classification score of action $\textit i$ output by the linear layer.
\vspace{-0.5cm}
\section{Experiments}
\subsection{Datasets and Implementation Details}
\textbf{Datasets:} \textbf{UAV-Human} \cite{Li_2021_CVPR} is a dataset designed for action recognition, comprising 22,476 video clips across 155 distinct classes. Crafted by a UAV traversing diverse urban and rural landscapes day and night, this dataset captures the essence of human activity. Curated from 119 unique individuals engaging in 155 varied activities across 45 environmental settings, it offers a rich description of human behavior. The original paper suggests two benchmark scenarios (CSv1 and CSv2) for evaluation: utilizing 89 subjects for training and 30 for testing, ensuring robustness and reliability in analysis.\\
\textbf{Implementation Details:} All experiments are conducted on the GeForce RTX 3090 GPUs. In the MixGCN branch, we employ TD-GCN \cite{TDGCN}, CTR-GCN \cite{CTR-GCN2021} and MST-GCN \cite{yan2018spatial} as the foundational components to derive classification scores based on GCN. Model training is executed using SGD over 65 epochs with a batch size of 64. Initially, the learning rate is established at 0.1 and undergoes decay by a factor of 0.1 at epochs 35 and 55, ensuring optimal convergence and performance refinement. Within the MixFormer branch, we integrate Skeleton MixFormer \cite{xin2023skeleton} as the core framework to derive classification scores based on Transformers. Employing SGD, we train the model across 90 epochs, utilizing a batch size of 128. The initial learning rate is defined at 0.02, ensuring a robust foundation for effective training and refinement.

\subsection{Comparisons with the State-of-the-Art}
In Table \ref{tab:sota}, we report the Top-1 accuracy of our HDBN and compare with the existing methods on the UAV-Human dataset. In this action recognition dataset, our HDBN outperforms most existing methods under the recommended evaluation benchmarks. In the UAV-Human dataset, the classification accuracy is 47.95$\%$ and 75.36$\%$ on the benchmark of CSv1 and CSv2 respectively, which outperforms CTR-GCN by 2.35$\%$ and 2.86$\%$. We also present the confusion matrices of our proposed HDBN in Fig. \ref{confusion}.

\subsection{Ablation Studies}
In Table \ref{tab:CTR-GCN-2D}, when using CTR-GCN as the backbone in the MixGCN branch, we present the recognition accuracy of four skeleton modalities in UAV-Human dataset. Meanwhile, in Table \ref{tab:CTR-GCN-3D}, we demonstrate the accuracy of estimating 3D poses from 2D poses using MotionBert and utilizing these 3D poses for action recognition. In addition, in Table \ref{tab:TD-GCN-2D} and Table \ref{tab:MST-GCN-2D}, we also report the recognition accuracy of using TD-GCN and MST-GCN to model 2D skeleton modalities. Finally, in Table \ref{tab:Mixformer-2D}, we present the recognition accuracy when using Skeleton MixFormer as the backbone in the MixFormer branch.

\vspace{-0.3cm}
\section{Conclusions}
We propose a novel dual-branch framework called Hybrid Dual-branch Network (HDBN) for robust skeleton-based action recognition, which introduces the Transformers and GCNs to model 2D and 3D skeleton data. Different from previous methods using single backbone or using different backbone to process various modalities, our HDBN is to leverage both Transformers and GCNs with the aim of robust skeleton-based action recognition. The effectiveness of HDBN is verified on the UAV-Human dataset, where our HDBN outperforms most existing methods.

\vspace{-0.3cm}
\section{Acknowledgement}
This work was supported by National Natural Science Foundation of China (No. 62203476), Natural Science Foundation of Shenzhen (No. JCYJ20230807120801002).

\small
\bibliographystyle{IEEEbib}
\bibliography{icme2023template}

\end{document}